# Deep reinforcement learning for time series: playing idealized trading games[*]

Xiang Gao[†]

Georgia Institute of Technology, Atlanta, GA 30332, USA

**Abstract**

Deep Q-learning is investigated as an end-to-end solution to estimate the optimal strategies for acting on time series input. Experiments are conducted on two idealized trading games. 1) Univariate: the only input is a wave-like price time series, and 2) Bivariate: the input includes a random stepwise price time series and a noisy signal time series, which is positively correlated with future price changes. The Univariate game tests whether the agent can capture the underlying dynamics, and the Bivariate game tests whether the agent can utilize the hidden relation among the inputs. Stacked Gated Recurrent Unit (GRU), Long Short-Term Memory (LSTM) units, Convolutional Neural Network (CNN), and multi-layer perceptron (MLP) are used to model Q values. For both games, all agents successfully find a profitable strategy. The GRU-based agents show best overall performance in the Univariate game, while the MLP-based agents outperform others in the Bivariate game.

## 1. Introduction

Learning to act optimally on time series input is of many practical uses in finance, healthcare, and industry. The value of taking an action depends on future actions and states, which makes it difficult to be modeled using a conventional supervised learning method. This is where reinforcement learning fits. Mnih *et al*. [1] has shown impressive results using such technique to let agent learn to play Atari games using deep Q-learning with experience replay. The model is based on convolutional neural network (CNN) and only inputs screenshots. No hand-crafted features are needed. Later the success of reinforcement learning is further demonstrated with AlphaGo Zero by Silver *et al*. [2]. It is promising to apply deep reinforcement learning to tasks where agents act on time series.

For sequence and time series modeling, recurrent neural network (RNN) is the one of the state-of-the-art model. It stores information over extended time intervals. Long Short-Term Memory (LSTM) [3, 4] is one popular architecture initially proposed to resolve the decaying error backflow issue. Gers *et al*. [5] showed that LSTM does not outperform for certain simpler time series and suggested to use LSTM only when simpler traditional approaches, such as multi-layer perceptron (MLP), fail. Malhotra *et al*. [6] used stacked LSTM networks for anomaly detection in time series.

---

They trained the model on normal time series, and then the prediction errors of this trained model is used to identify abnormal behavior. This approach is helpful as in the real world instances of normal behavior is usually abundant but instances of anomalous are rare. The model was tested on four real-world datasets, and showed that better or similar performance comparing to a stacked RNN using sigmoid units. Lipton *et al.* [7] used LSTM to classify multivariate time series of clinical measurements, and outperformed several baselines including a MLP model trained on hand-engineered features. Gated Recurrent Unit (GRU) [8] is another popular RNN architecture that is simpler comparing to LSTM. Empirical work has shown that the performance of GRU is comparable with LSTM on sequence modeling [9, 10]. Recently more RNN architectures are proposed, including gated-feedback recurrent neural network [11], recurrent latent variable model [12], and hierarchical multiscale RNN [13]. Karpathy *et al.* [14] visualized RNN and provided an analysis using character-level languages. They found the existence of interpretable cells keeping long-range dependencies such as quotes, brackets, and line-lengths.

CNN also shows its potential in time series application, besides its successful application in image recognition [15, 16]. Zheng *et al.* [17] proposed a CNN based method for multivariate time series classification. For each channel, a stacked CNN is used to extract features, then the features for all channels join to a MLP which finally outputs the clarification. The proposed method was tested on two multivariate datasets and outperformed the benchmark methods, 1-nearest neighbor and MLP. Further improvement was found using a deeper architecture. Recognizing the fact that time series often have features at different time scales, Cui *et al.* [18] transform the input time series to several branches, including the original branch, multi-scale branch by down-sampling at different rate, and multi-frequency branch by moving average. Each branch is sent to a CNN and then joining in a global CNN, after which a MLP is used for final classification output. This method is tested and compared with various benchmark methods. Better accuracy of the proposed method is shown. Borovykh et al. [19] proposed a method to forecast multivariate time series based on deep convolutional WaveNet architecture. The network is based on stacked dilated convolutions, which allows a broad receptive field. The method is tested to forecast index and foreign exchange rates, and showed smaller error comparing to autoregressive model and a LSTM network. Wang *et al.* [20] proposed a fully convolutional network as a baseline for time series classification. The model is tested on univariate time series and compared with a MLP model, a residual network, and eight other benchmark methods. The fully convolutional network showed the best performance.

As both RNN and CNN have been used in sequence modeling as introduced above, Yin *et al.* [21] compared RNN and CNN regarding their performance for natural language processing. They found that RNN performs better when the global/long-range semantics is required, and CNN is better when local key-phrases are important.

For many time series, especially these in financial markets, it is generally difficult for human investigators to make prediction accurately, thus providing high-quality training and test datasets. Therefore, as a starting point, in the present study, artificially generated inputs are used. They are designed to test if the trained agent can capture some simple dynamics and hidden relation among inputs. Stacked GRU, LSTM, CNN and MLP are used to model Q values.

## 2. Games

*2.1 Overview*

When playing games, agent learns to trade a hypothesized stock and make profit by buying at low and selling at high price. At each time step, the agent is given observations in the past 40 time steps and then it chooses an action (see Section 2.2), which results in a reward (see Section 2.3). After $T$=180 time steps, an episode ends. Two games are considered:

- *Univariate*: the only input is an artificially generated wave-like price time series. The shape of the input is $T \times 1$. Some examples are shown in Figure 1. The time series is generated from superposition of short-term waves, a long term wave and noise. The short-term waves are sine functions with random periods (range 10~40) and random amplitudes (range 5~80). And the long-term wave is also a sine function, with a random period (range 80~200) and a random amplitude (range 20~80)

- *Bivariate*: the inputs are two time series, one is a random stepwise price time series and another is a noisy leading signal time series. The movement of the signal is positively correlated with future price changes with a random forecast horizon. The shape of the input is $T \times 2$. Some examples are shown in Figure 2. At a random number of time steps (range 15~30), the price jumps for a random amplitude (-30~30). The signal is generated by moving the price ahead by a random number of time steps (range 10~30) and then adding a noise term.

The Univariate game is to test if the model can capture the underlying dynamics, and the Bivariate game task is to test if the model can utilize the hidden relation among the inputs. For both games, the inputs are positive values.

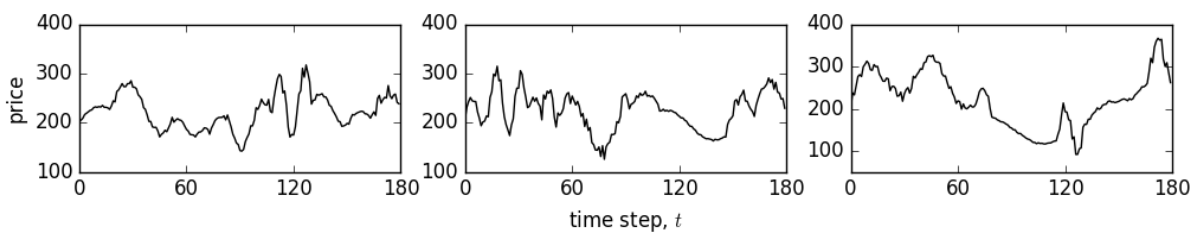

Figure 1 Examples of the Univariate game input

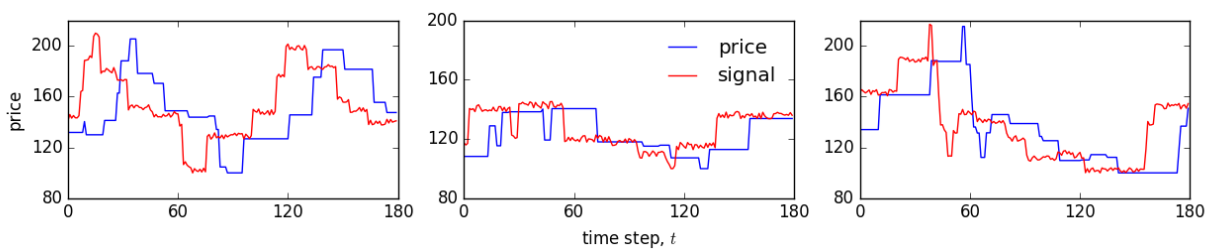

Figure 2 Examples of the Bivariate game input

*2.2 Actions*

Three actions are considered: CASH (don not buy stock, or sell existing stock), BUY (buy a share of stock), HOLD (hold the bought stock). Please note that BUY differs from HOLD as a cost is applied for BUY, but not applied for HOLD. Depending on whether a share of stock is already bought, valid actions are different:

$$\text{valid actions} = \begin{cases} \{\text{CASH, BUY}\} & \text{not holding any stock} \\ \{\text{CASH, HOLD}\} & \text{otherwise} \end{cases}$$

*2.3 Reward*

The reward for the agent to take action at time step t is

$$r_t = \begin{cases} 0 & \text{CASH} \\ p_{t+1} - p_t - c & \text{BUY} \\ p_{t+1} - p_t & \text{HOLD} \end{cases}$$

where $p$ is price and constant $c = 3.3$ is the cost to buy stock.

## 3. Methodology

*3.1 Reinforcement learning*

Following Ref. [1], deep Q-learning with experience replay is applied here. The experience of the agent is stored and randomly replayed at each time step for training. The input state $s_t$ (a clip of time series) is mean-value-normalized. The value of Q for a given action and state is the estimated expectation of total future rewards (profit and loss for the present work) discounted at the current step. The algorithm is presented in Table 1.

Table 1 Deep Q-learning algorithm adapted from Ref. [1]

**For** episode = 1 to $N$
    **For** time step $t$ = 1 to $T$
        With probability $\epsilon$ select a random valid action $a_t$
        Otherwise select the valid action $a_t$ that maximize predicted Q
        Given $a_t$, emulator returns reward $r_t$ and new state $s_{t+1}$
        Store ($s_t$, $a_t$, $r_t$, $s_{t+1}$) in memory
        **For** memory ($s_j$, $a_j$, $r_j$, $s_{j+1}$) in sampled minibatch
            perform gradient descent on $Q(s_j, a_j)$ using target value
$$Q_{\text{target}}(s_j, a_j) = \begin{cases} r_j & \text{game end at } j \\ r_j + \gamma \max_{\text{valid } a} Q(s_{j+1}, a) & \text{otherwise} \end{cases}$$
        **end for**
    **end for**
**end for**

*3.2 Model architecture*

Although more sophisticated architectures exist, the present work focuses on the performance of some baseline models. Three neural networks are considered to estimate Q value. These networks have the same input (with shape of $T \times 1$ or $T \times 2$, depending on the game as introduced in Section 2.1) and the same output shape of $3 \times 1$, which corresponds to the predicted Q value of three actions. For all of the them, the output layer is a fully connected layer of 3 units with linear activation function. They differ from each other on the hidden layers, as introduced below.

- CNN: the net is named as CNN-*F*x*L*, where *F* is the number of filters in a convolutional layer and *L* is the number of convolutional layer. The length of the filter is fixed as 3 and the stride is fixed as 1. Each convolutional layer is applied with a rectifier nonlinearity, before being connected to a max-pooling layer of pooling size 2. These stacked layers are followed by two fully-connected layers with rectifier nonlinearity, the first one has 48 units and the second has 24 units.

- GRU: the net is named as GRU-*C*x*L*, where *C* is the number of output channels from a GRU layer layer and *L* is the number of stacked GRU layers. The last GRU layer only output the values at the last time step (i.e. shape of *1*x*C*), while the GRU layers before that output sequences (i.e., shape of 40x*C*). The last GRU unit is followed by two fully-connected layers with rectifier nonlinearity, both has *C* units.

- LSTM: the architecture is same to the GRU-based, except the GRU unit is replaced by LSTM unit.

- MLP: the net is named as MLP-*H*x*L*, where *H* is the number of hidden units in each fully connected layer and *L* is the number of fully connected layers.

## 4. Experiments

The agents are trained with the Q-learning algorithm introduced in Section 3.1 by playing 1000 episodes of games with a discount factor $\gamma$ of 0.8, and then tested by playing 100 episodes of games. All agents use the same training and test dataset. For each type of model considered, four architectures of different depth and width are considered. The test results are shown in Table 2. For the Univariate game, the GRU-based agents significantly outperformed the agents based on other models, especially in terms of mean profit and loss (P&L). For the Bivariate game, the winner is MLP-based agents. CNN-based agents again showed worst performance for both games. The number of layers and hidden units do affect the performance. However the relation is generally not monotonic.

An example of the episode played by the GRU-agent during test is illustrated in Figure 3. The agent realized that the Q-value corresponding to BUY is lower than HOLD (because of the cost), and the fact that the Q-value corresponding to CASH should be always greater than zero. Although not shown, this is also true for other models and for the Bivariate game.

Table 2 Test results for the profit and loss (P&L) generated per episode

|  | Univariate game | | | | | Bivariate game | | | |
|---|---|---|---|---|---|---|---|---|---|
|  | | In-sample | | Out-of-sample | | | In-sample | | Out-of-sample | |
|  | #param | Mean P&L | P&L>0 | Mean P&L | P&L>0 | #param | Mean P&L | P&L>0 | Mean P&L | P&L>0 |
| MLP 16x4 | 1,523 | 104.3 | 100% | 96.9 | 97% | 2,163 | 49.4 | 92% | 40.5 | 96% |
| MLP 16x5 | 1,795 | 103.2 | 100% | 91.7 | 100% | 2,435 | 47.4 | 94% | 43.8 | 94% |
| MLP-32x4 | 4,579 | 122.4 | 100% | 86.1 | 96% | 5,859 | 49.3 | 92% | 37.4 | 95% |
| MLP-32x5 | 5,635 | 125.0 | 100% | 89.8 | 99% | 6,915 | 52.7 | 95% | 42.0 | 96% |
| GRU-8x3 | 1,227 | 116.1 | 100% | 115.6 | 100% | 1,251 | 35.0 | 81% | 31.8 | 86% |
| GRU-16x3 | 4,627 | 114.0 | 100% | 112.1 | 100% | 4,675 | 44.0 | 84% | 35.0 | 91% |
| GRU-16x2 | 3,043 | 120.7 | 99% | 115.2 | 100% | 3,091 | 46.5 | 91% | 31.1 | 88% |
| GRU-32x3 | 17,955 | 143.3 | 100% | 116.1 | 99% | 18,051 | 51.5 | 94% | 37.2 | 94% |
| LSTM-8x3 | 1,579 | 58.9 | 90% | 68.0 | 94% | 1,611 | 37.6 | 80% | 32.5 | 84% |
| LSTM-16x3 | 5,971 | 84.8 | 98% | 79.4 | 98% | 6,035 | 44.5 | 91% | 38.4 | 94% |
| LSTM-16x2 | 3,859 | 106.3 | 100% | 107.8 | 100% | 3,923 | 44.0 | 87% | 37.8 | 90% |
| LSTM-32x3 | 23,203 | 134.1 | 100% | 98.5 | 97% | 23,331 | 44.3 | 90% | 31.0 | 89% |
| CNN-8x3 | 2,883 | 89.0 | 98% | 77.3 | 94% | 2,907 | 39.1 | 86% | 34.7 | 87% |
| CNN-16x3 | 5,235 | 106.0 | 98% | 86.7 | 95% | 5,283 | 38.8 | 85% | 28.5 | 85% |
| CNN-16x2 | 8,291 | 108.5 | 100% | 77.1 | 95% | 8,339 | 46.4 | 91% | 33.8 | 89% |
| CNN-32x3 | 12,243 | 120 | 100% | 80.8 | 96% | 12,339 | 32.3 | 85% | 15.4 | 72% |

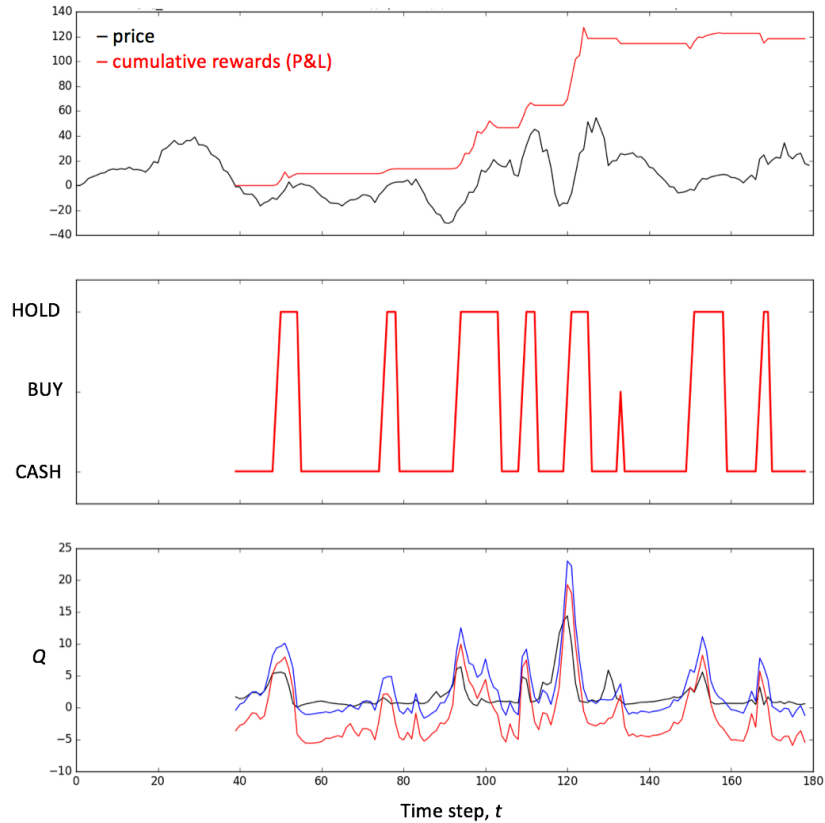

Figure 3 A test episode of the Univariate game played by the trained GRU-based agent

# 5. Conclusion and future works

The demonstration given in the present work shows that deep reinforcement learning is promising to discover the optimal strategies to act on time series input. The agent is tested whether they can capture the underlying dynamics (the Univariate game) and utilize the hidden relation among the inputs (the Bivariate game). The agents based on GRU showed best overall performance in the Univariate game and slightly underperformed the agents based on MLP, but was still able to find profitable strategies. In the future, more advanced reinforcement learning technique will be employed and the agents will be tested in more realistic games.

17. Zheng, Y., et al. *Time series classification using multi-channels deep convolutional neural networks*. in *International Conference on Web-Age Information Management*. 2014. Springer.
18. Cui, Z., W. Chen, and Y. Chen, *Multi-scale convolutional neural networks for time series classification.* arXiv preprint arXiv:1603.06995, 2016.
19. Borovykh, A., S. Bohte, and C.W. Oosterlee, *Conditional time series forecasting with convolutional neural networks.* arXiv preprint arXiv:1703.04691, 2017.
20. Wang, Z., W. Yan, and T. Oates. *Time series classification from scratch with deep neural networks: A strong baseline*. in *Neural Networks (IJCNN), 2017 International Joint Conference on*. 2017. IEEE.
21. Yin, W., et al., *Comparative study of cnn and rnn for natural language processing.* arXiv preprint arXiv:1702.01923, 2017.